\documentclass{article}

\usepackage{microtype}
\usepackage{graphicx}
\usepackage{booktabs} %

\usepackage{hyperref}

\usepackage[accepted]{icml2021}

\usepackage{url}
\usepackage{amsfonts}       %
\usepackage{nicefrac}       %
\usepackage{amsfonts, amsmath}
\usepackage{tikz}
\usepackage{pgf}
\usepackage{enumitem}
\usepackage{mathtools}
\usepackage{bbm}
\usepackage{enumitem}
\usepackage{amsthm,amssymb}
\usepackage{amssymb}
\usepackage{booktabs}
\usepackage{float}
\usepackage{wrapfig}
\usepackage{caption} 
\usepackage{xspace}
\usepackage{xcolor}
\usepackage{caption}
\usepackage{subcaption}
\usepackage[capitalise,noabbrev,nameinlink]{cleveref}

\usepackage{multirow}

\usepackage{amsmath,amsfonts,bm}

\def\eqref#1{equation~\ref{#1}}

\def\1{\bm{1}}

\DeclareMathAlphabet{\mathsfit}{\encodingdefault}{\sfdefault}{m}{sl}
\SetMathAlphabet{\mathsfit}{bold}{\encodingdefault}{\sfdefault}{bx}{n}

\newcommand{\E}{\mathbb{E}}

\creflabelformat{equation}{#2\textup{#1}#3}
\crefname{algocf}{Algorithm}{Algorithms}
\Crefname{algocf}{Algorithm}{Algorithms}

\newcommand{\changed}[1]{{\color{olive}#1}}

\renewcommand{\changed}[1]{#1}

\newcommand{\J}{\text{J}}
\newcommand{\JT}{\text{J}^\text{T}}
\newcommand{\pluseq}{\mathrel{+}=}

\newcommand{\lat}{z}
\newcommand{\obs}{o}

\usepackage{tikz}
\def\checkmark{\tikz\fill[scale=0.4](0,.35) -- (.25,0) -- (1,.7) -- (.25,.15) -- cycle;}

\begin{document}

\twocolumn[
\icmltitle{Model-Based Reinforcement Learning via Latent-Space Collocation}

\icmlsetsymbol{equal}{*}

\begin{icmlauthorlist}
\icmlauthor{Oleh Rybkin}{equal,penn}
\icmlauthor{Chuning Zhu}{equal,penn}
\icmlauthor{Anusha Nagabandi}{cov}
\icmlauthor{Kostas Daniilidis}{penn}
\icmlauthor{Igor Mordatch}{goog}
\icmlauthor{Sergey Levine}{ucb}
\end{icmlauthorlist}

\icmlaffiliation{penn}{University of Pennsylvania}
\icmlaffiliation{ucb}{UC Berkeley}
\icmlaffiliation{cov}{Covariant}
\icmlaffiliation{goog}{Google AI}
\icmlcorrespondingauthor{Oleh Rybkin}{oleh@seas.upenn.edu}

\icmlkeywords{Machine Learning, ICML}

\vskip 0.3in
]

\printAffiliationsAndNotice{\icmlEqualContribution} %

\begin{abstract}

The ability to plan into the future while utilizing only raw high-dimensional observations, such as images, can provide autonomous agents with broad capabilities. 
Visual model-based reinforcement learning (RL) methods that plan future actions directly have shown impressive results on tasks that require only short-horizon reasoning, however, these methods struggle on temporally extended tasks. We argue that it is easier to solve long-horizon tasks by planning sequences of states rather than just actions, as the effects of actions greatly compound over time and are harder to optimize.
To achieve this, we draw on the idea of collocation, which has shown good results on long-horizon tasks in optimal control literature, and adapt it to the image-based setting by utilizing learned latent state space models.
The resulting latent collocation method (LatCo) optimizes trajectories of latent states, which improves over previously proposed shooting methods for visual model-based RL on tasks with sparse rewards and long-term goals. Videos and code: \url{https://orybkin.github.io/latco/}.

\end{abstract}

\section{Introduction}

\begin{figure}%
    \centering
    \includegraphics[width=\linewidth]{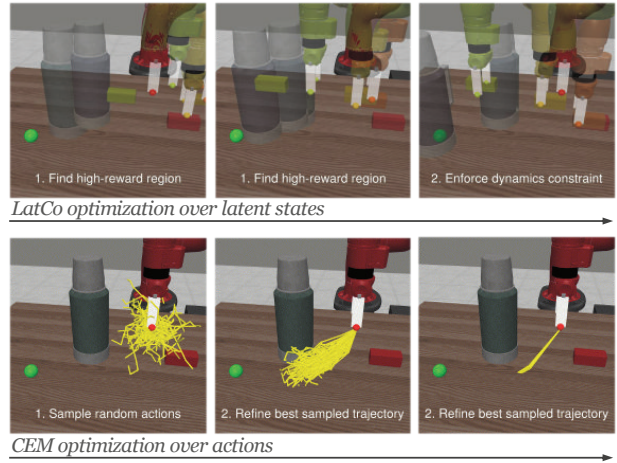}
    \vspace{-0.2in}
    \caption{\textbf{Top}: Latent collocation (LatCo) on a tool use task, where the thermos needs to be pushed with the stick. Each image shows a full plan at that optimization step, visualized via a diagnostic network. LatCo optimizes a latent state sequence and is able to temporarily violate the dynamics during planning, such as the stick flying in the air without an apparent cause. This allows it to rapidly discover the high-reward regions, while the subsequent refinement of the planned trajectory focuses on feasibly achieving it. \textbf{Bottom}: in contrast, shooting optimizes an action sequence directly and is unable to discover picking the stick as the actions that lead to that are unlikely to be sampled.
    }
\label{fig:teaser}
\vspace{-0.2in}
\end{figure}

For autonomous agents to perform complex tasks in open-world settings, they must be able to process high-dimensional sensory inputs, such as images, and reason over long horizons about the potential effects of their actions. Recent work in model-based reinforcement learning (RL) has shown impressive results in autonomous skill acquisition directly from image inputs, demonstrating improvements both in terms of data efficiency and generalization~\citep{ebert2018visual,hafner2018learning,zhang2019solar,schrittwieser2019mastering}.
While these advancements have been largely fueled by improvements in modeling \citep{finn2016unsupervised,hafner2018learning}, 
they leave much room for improvement in terms of planning and optimization. Many of the current best-performing deep model-based RL approaches use only gradient-free action sampling as the underlying optimizer~\citep{chua2018deep,nagabandi2020deep}, and are typically applied to tasks with very short horizons, such as planning a sequence of 5 \citep{ebert2018visual} or 12 \citep{hafner2018learning} actions. {On longer-horizon tasks, we observe that these shooting methods struggle with local optima, as credit assignment for individual actions becomes harder}. In this work, we study how more powerful planners can be used with these models to achieve long-horizon reasoning.

It is appealing instead of optimizing a sequence of actions to optimize a sequence of states.
While small deviations in actions can greatly compound over time and affect the entire trajectory, states can often be easily optimized locally just based on the neighboring states in the trajectory (see \cref{fig:teaser}). An approach that optimizes states directly could then perform credit assignment more easily and be better conditioned. 
However, it is necessary to ensure that the optimized state sequence is dynamically feasible -- that is, each state in the trajectory is reachable from the previous state. Prior deep reinforcement learning methods have addressed this problem by learning reachability functions and performing graph search on the replay buffer \cite{savinov2018semi,kurutach2018learning,eysenbach2019search}.
However, it is unclear how to use these methods with partial observability, stochastic dynamics, or reward functions beyond goal reaching. Instead, to arrive at an optimal control solution,
we turn to the technique of collocation,
which optimizes a sequence of states to maximize the reward, while also eventually ensuring dynamics feasibility by recovering the corresponding actions in a constrained optimization problem:
\begin{equation}
    \label{eq:colloc}
    \max_{s_{2:T}, a_{1:T-1}} \sum_t r(s_t) \quad \text{s.t.} \quad s_{t+1} = f(s_t,a_t).
\end{equation}
Collocation only requires learning a dynamics model and a reward function, and can be used as a plug-and-play optimizer within common model-based reinforcement learning approaches, while providing a theoretically appealing formulation for optimizing sequences of states. 

Collocation, as introduced above, can provide many benefits over other optimization techniques, but it has thus far been demonstrated~\citep{liu2002synthesis,ratliff2009chomp,schulman2014motion,kalakrishnan2011stomp,posa2014direct} mostly in conjunction with known dynamics models or when performing optimization over state vectors with a few tens of dimensions \cite{bansal2016learning,du2019model}.
In this work, we are interested in autonomous behavior acquisition directly from image inputs, where both the underlying states as well as the underlying dynamics are unknown, and the dimensionality of the observations is in the thousands. Na\"{i}vely applying collocation to sequences of images would lead to intractable optimization problems, due to the high-dimensional as well as partially observed nature of images.
Instead, we draw on the representation learning literature and leverage latent dynamics models, which learn a latent representation of the observations that is not only Markovian but also compact,
and lends itself well to planning. In this learned latent space, we propose to perform collocation over states and actions with the joint objective of maximizing rewards as well as ensuring dynamic feasibility.

The main contribution of this work is an algorithm for latent-space collocation (LatCo), which provides a practical way to utilize collocation methods within a model-based RL algorithm with image observations. LatCo plans sequences of latent states directly from image observations, and is able to plan under uncertainty. 
To the best of our knowledge, our paper is the first to scale collocation to visual observations,
enabling longer-horizon reasoning by drawing both on techniques from trajectory optimization and deep visual model-based RL. We show experimentally that our approach achieves strong performance 
on challenging long-horizon visual robotic manipulation tasks where prior shooting-based approaches fail.

\section{Related Work}

\begin{figure*}
    \includegraphics[width=\linewidth]{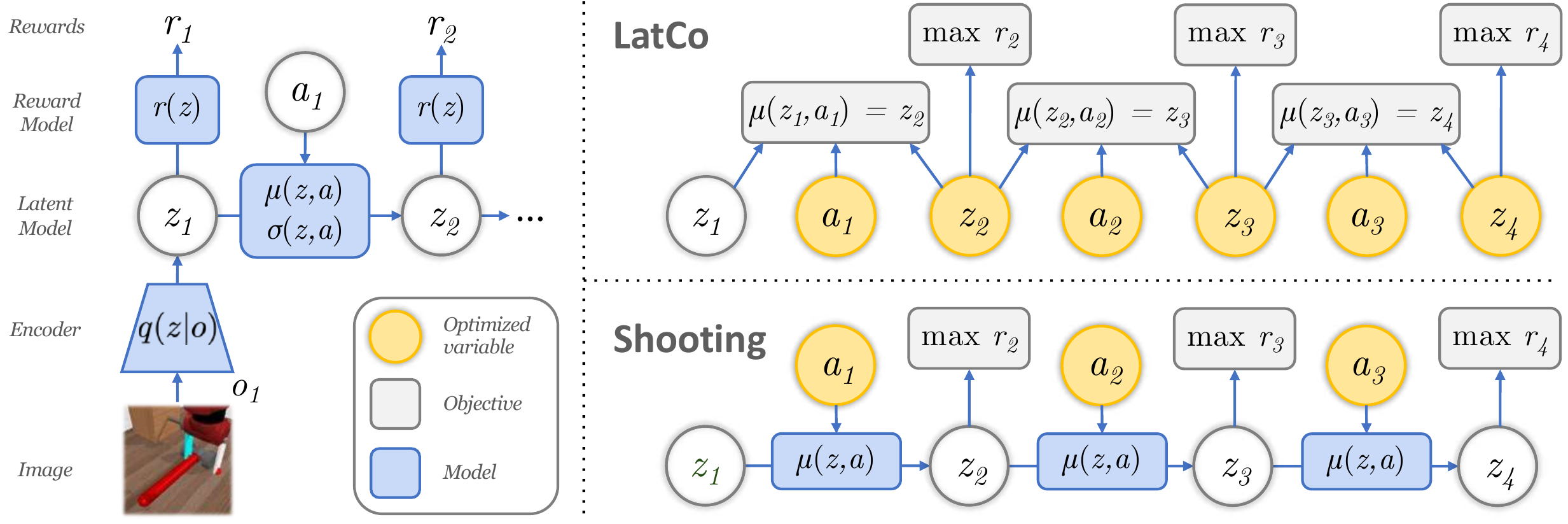}   
    \vspace{-0.15in}
    \caption{ 
    Latent Collocation (LatCo). \textbf{Left}: Our latent state-space model, with an encoder $q(z|o)$ and a latent state-space dynamics model {$p(\lat_{t+1}|\lat_t,a_t) \sim \mathcal{N}(\mu(\lat_t, a_t), \sigma(\lat_t, a_t))$}. A reward model $r(\lat_t)$ predicts the reward from the latent state. {The model is trained with a variational lower bound to reconstruct the observations (not shown)}. \textbf{Right}: comparison of deterministic LatCo and shooting methods. LatCo optimizes a sequence of latent states and actions $\lat_{2:T},a_{1:T}$ to maximize rewards $r(\lat_t)$ as well as satisfy dynamics $\lat_{t+1} = \mu(\lat_t, a_t)$. This joint optimization allows  \changed{the dynamics constraint to be relaxed at first,} which helps escape local minima. In contrast, shooting methods require recursive application of the dynamics and backpropagation through time, which is often difficult to optimize.
    }
    \label{fig:method}
    \vspace{-0.1in}
\end{figure*}

\textbf{Planning in model-based RL.}
Many recent papers on deep model-based RL
\citep{chua2018deep,ebert2018visual} use the cross-entropy method \citep{blossom2006cross} to optimize action trajectories in a shooting formulation. Other work has explored  different optimization methods such as the iterative linear-quadratic regulator \citep{levine2013guided,watter2015embed,zhang2019solar},
or Monte-Carlo tree search \citep{schrittwieser2019mastering}.
However, these shooting approaches rely on local search in the space of actions, which is prone to local optima. 
Instead, our collocation approach optimizes over (latent) states as opposed to actions, which we show often enables us to escape local minima and plan better.
Another line of work, inspired by classical sampling-based planning \cite{kavraki1996probabilistic,lavalle1998rapidly}, uses graph-based optimization \citep{kurutach2018learning,savinov2018semi,eysenbach2019search,liu2020hallucinative} {or other symbolic planners \citep{asai2017classical}} to optimize a sequence of states. While this escapes the local minima problem of shooting methods, such graph-based methods require constructing a large graph of possible states and scale poorly to combinatorial state spaces
and context-dependent tasks. In contrast, latent collocation provides a principled control method that is able to optimize a sequence of states using continuous optimization, and does not suffer from the drawbacks of graph-based methods.

Recent work has designed hierarchical methods that plan over extended periods of time with intermediate subgoals, and then use separate model-free \citep{pong2018temporal,nasiriany2019planning} or model-based \citep{pertsch2019keyin,nair2019hierarchical,pertch2020long,parascandolo2020divide} controllers to reach the subgoals. This can be considered a hierarchical form of collocation-based planning. However, in contrast to these approaches, which require a separate controller for reaching subgoals, we focus on the standard model-based RL setup where only a latent dynamics model is learned, and show that latent collocation performs long-horizon tasks without hierarchical planning. We further discuss state optimization objectives in \cref{app:objectives}.

Many of the previously proposed model-based methods~\citep{wahlstrom2015pixels,watter2015embed,kurutach2018learning,buesing2018learning,zhang2019solar,hafner2018learning}
use latent state-space models for improved prediction quality and runtime. Our proposed method leverages this latent state-space design to construct an effective trajectory optimization method with collocation, {and we design our method to be model-agnostic, such that it can benefit from future improved latent variable models}.

\textbf{Collocation-based planning.
\footnote{In this paper we use the terms ``trajectory optimization" and ``planning" synonymously as is common in MBRL literature.}
}
Collocation is a powerful technique for trajectory optimization \citep{hargraves1987direct,witkin1988spacetime,betts1998survey,tedrake2021notes} that optimizes a sequence of states for the sum of expected reward, while eventually enforcing the constraint that the optimized trajectory conform to a dynamics model for some actions (also see \citet{kelly2017introduction} for a recent tutorial). Prior work in optimal control and robotics has explored many variants of this approach, with Hamiltonian optimization \citep{ratliff2009chomp}, explicit handling of contacts \citep{mordatch2012discovery,posa2014direct}, sequential convex programming \citep{schulman2014motion}, {as well as stochastic trajectory optimization \citep{kalakrishnan2011stomp}}. \citet{platt2010belief,patil2015scaling} proposed probabilistic extensions of collocation. These works have demonstrated good results in controlling complex simulated characters, such as humanoid robots, contact-heavy tasks, and tasks with complex constraints. {
The optimization algorithm we use is most similar to that of \citet{schulman2014motion},} however, all this prior work assumed availability of a ground truth model of the environment and low-dimensional state descriptors. 

Some recent works have attempted using collocation with learned neural network dynamics models \citep{bansal2016learning,du2019model}, but only considered simple or
low-dimensional dynamics. In this work, we address how to scale up collocation methods to high-dimensional image observations, where direct optimization over images is intractable, and the dynamics are more challenging to learn. We propose to do this by utilizing a learned latent space.
\section{Latent Collocation (LatCo)} 

We aim to design a collocation method that plans trajectories from raw image observations. A na\"{i}ve approach would learn an image dynamics model, and directly optimize an image sequence using \cref{eq:colloc}. 
However, this is impractical for several reasons. First, optimizing over images directly is difficult due to the high dimensionality of the images and the fact that valid images lie on a thin manifold. 
Second, images typically do not constitute a Markovian state space. We propose to instead learn a Markovian and compact space by means of a latent variable model, and then use this learned latent space for collocation.

\subsection{Learning Latent Dynamics}

\label{sec:back_models}

The design of dynamics models for visual observations is challenging. Recent work has proposed learning latent state-space models that represent the observations in a compact latent space $z_t$. Specifically, this work learns a latent dynamics model $p_\phi(\lat_{t+1}|\lat_t, a_t)$, as well as observation $p_\phi(\obs_t|\lat_t)$ and reward $r(\lat_t)$ decoders (see \cref{fig:method}, left).  This approach is powerful due to high-capacity neural network latent dynamics models, and is computationally efficient as the latent space is compact. Importantly, the Markov property of the latent state is enforced, allowing a convenient interpretation of the latent dynamics model as a Markov decision process. The model is shown in \cref{fig:method} (left).

Many environments of interest are stochastic or partially observable, which necessitates accounting for uncertainty. The latent dynamics distribution $p_\phi(\lat_{t+1}|\lat_t, a_t)$ should then reflect the stochasticity in the observed data.
To achieve this, we train the model by
maximizing the likelihood of the observed data $r_{1:T}, o_{1:T}$. While maximizing exact likelihood is often intractable, we optimize a variational lower bound on it using a variational distribution $q_\phi(\lat_{t+1} | \obs_{t+1}, a_{t}, \lat_t)$ \citep{chung2015recurrent,fraccaro2016sequential}:
\begin{multline}
    \label{eq:elbo} 
     \ln p_\phi(o_{2:T},r_{1:T}|o_1,a_{1:T}) \geq \mathcal{L}_{\text{ELBO}}(o_{1:T},a_{1:T},r_{1:T}) =  \\ 
     \E_{q_\phi(\lat_{1:T} | \obs_{1:T}, a_{1:T}, \lat_0)} \sum_t \big[\ln p_\phi(\obs_{t+1}, r_{t+1} \vert \lat_{t+1}) - \\
     \text{KL}\left(q_\phi(\lat_{t+1} | \obs_{t+1}, a_{t}, \lat_t) \;\vert\vert\; p_\phi(\lat_{t+1} | \lat_{t}, a_t)\right) \big].
\end{multline}

\subsection{Latent Collocation with Probabilistic Dynamics}
\label{sec:back_colloc}

Our probabilistic latent-space dynamics learns a latent MDP $p_\phi(z_{t+1} | z_t, a_t), r(z_t)$, which can be unrolled into the future without decoding the observations. Given the current state $z_1$, we can therefore select the actions with maximum expected return by planning in this latent MDP: 
\begin{equation}
    \max_{a_{1:T-1}} \E_{p_\phi(z_{t+1} | z_t, a_t)}\left[\sum_t r(z_t)\right].
    \label{eq:trajopt}
\end{equation}
Prior work \cite{ebert2018visual,hafner2018learning} used shooting methods \citep{betts1998survey,tedrake2021notes} to directly
optimize actions with this objective.
However, this is known to be poorly conditioned due to the recursive application of the dynamics, which results in hard credit assignment.
Instead, LatCo leverages the structure of the problem to construct an objective with only pairwise dependencies between temporally adjacent latents, and no recursive application. 

We would like to formulate the trajectory optimization problem in \cref{eq:trajopt} as a constrained optimization problem, optimizing over sequences of both latent states and actions. However, this requires reformulating the problem in terms of probability distributions, since the model has stochastic dynamics and each observation corresponds to a distribution over latent states. To handle this, we formulate collocation with distributional constraints, analogously to belief-space planinng~\citep{platt2010belief,patil2015scaling}.
This problem can be defined abstractly as optimizing a sequence of distributions $q(z_t)$, each representing an uncertain estimate of what the latent $z_t$ will be at time $t$:
\begin{align}
\begin{split}
    \label{eq:prob_colloc}
    \max_{q(z_{2:T}), a_{1:T-1}} & \sum_t \E_{q(z_t)} \left[r(z_t)\right] \\
    \text{s.t.} \quad & q(z_{t+1}) = \mathbb{E}_{q(z_t)} p_\phi(z_{t+1}|z_t,a_t). \\
    & q(z_{1}) = p(z_1).
\end{split}
\end{align}

When the constraint is satisfied, this is equivalent to the original problem in \cref{eq:trajopt}. A simplified version of this approach is illustrated in \cref{fig:method} (right). We can express the distributional constraint in the form of a Bregman divergence or moment matching. For computational simplicity, we follow the latter approach
\begin{align}
\begin{split}
    \label{eq:prob_colloc_moments}
    \max_{q(z_{2:T}), a_{1:T-1}} & \sum_t \E_{q(z_t)} \left[r(z_t)\right] \\
    \text{s.t.} \quad & \text{mean}[p(z_{t})] = \text{mean}[q(z_{t})]. \\
    & \text{var}[p(z_{t})] = \text{var}[q(z_{t})]. \\
\end{split}
\end{align}
where we denoted with some abuse of notation $p(z_{t+1}) = \mathbb{E}_{q(z_t)} p_\phi(z_{t+1}|z_t,a_t).$ Further moments can be used for distributions with more degrees of freedom.

\subsection{LatCo with Deterministic Plans}

In many cases, finding a good plan without accounting for stochasticity is an effective and computationally efficient strategy. To do this, we will represent the plan distribution $q(z_{2:T})$ with particles ${z_{2:T}^i}_i$. In practice, we will simply use a single particle. The moment matching constraints from \cref{eq:prob_lagrangian} therefore reduce to a single constraint:
$\text{mean}[p_\phi(\lat_{t} | \lat_{t-1}, a_{t-1})] = z_t$, 
while the variance constraint disappears since the variance of a set of one particle is a constant. This constraint, for $\mu_\phi(z_t,a_t) = \E_{p_\phi(\lat_{t+1} | \lat_t, a_t)} \left[ \lat_{t+1} \right]$, yields a simplified planning problem:
\begin{equation}
    \label{eq:deter_colloc}
    \max_{z_{2:T}, a_{1:T-1}} \sum_t r(z_t) \quad \text{s.t.} \quad z_{t+1} = \mu_\phi(z_t,a_t).
\end{equation}

This approximation is equivalent to assuming that the underlying dynamics model is deterministic using its expected value, and performing standard deterministic collocation. This is a common assumption known as the certainty equivalence principle, and has appealing properties for certain kinds of distributions \cite{aoki1967optimization,bar1974dual,kwakernaak1972linear}. Other approximations with a single particle are possible, such as maximizing the likelihood of the particle, but these require introduction of additional balance hyperparameters, while our constrained optimization approach is principled and automatically tunes this balance. We visualize this approach in \cref{fig:method}.

\subsection{LatCo with Gaussian Plans}
\label{sec:platco}

The simple deterministic approximation from the previous subsection can perform well in environments that are close to deterministic, however, since it approximates the distribution $q(z_t)$ with just a single particle it is not expressive enough to represent the uncertainty in the plan. Instead, we can use more expressive distributions in our LatCo framework to perform planning under uncertainty. Specifically, since our latent dynamics $p_\phi(z_{t+1} | z_t, a_t)$ follow a Gaussian distribution, we parametrize $q$ as a diagonal-covariance Gaussian $q(z_{1:T}) = \mathcal{N}(\mu_{1:T}, \sigma_{1:T}^2)$. More expressive distributions can be used, but we observed that diagonal-covariance Gaussian already yields good performance on stochastic environments. To evaluate the moment matching terms in \cref{eq:prob_colloc_moments}  we use $\mu, \sigma^2$ directly for the mean and variance of $q(z_t)$, while the mean and variance of $p(z_t)$ are estimated with samples. Gradients are estimated with reparametrization and 50 samples.

\subsection{LatCo for Visual MBRL}

We discussed two objectives that either represent the plan $q(z_t)$ deterministically with a single particle or with a Gaussian distribution. Either objective can be used in our framework for latent collocation. Under some regularity conditions, we can reformulate the equivalent dual version of \cref{eq:prob_colloc_moments} as \changed{the saddle point problem on the Lagrangian}
\begin{align}
\begin{split}
    \label{eq:prob_lagrangian}
    \min_\lambda & \max_{q(z_{2:T}), a_{1:T-1}} \sum_t \E_{q(z_t)} \left[r(z_t)\right] \\
    & \quad - \lambda  (||\text{mean}[p(z_{t})] - \text{mean}[q(z_{t})]||^2 - \epsilon) \\
    & \quad - \lambda  (||\text{var}[p(z_{t})] - \text{var}[q(z_{t})]||^2 - \epsilon),
\end{split}
\end{align}
\changed{by introducing Lagrange multipliers $\lambda$}. In our implementation, we used squared constraints $||\text{mean}[p(z_{t})] - \text{mean}[q(z_{t})]||^2 = \epsilon$, however a non-squared constraint can also be used. Similar to dual descent \cite{nocedal2006numerical}, we address this min-max problem by taking alternating maximization and minimization steps, as also used with deep neural networks by~\citet{goodfellow2014generative,haarnoja2018soft}. While this strategy is not guaranteed to converge to a saddle point, we found it to work well in practice. \changed{Following prior work, we do not learn a terminal value function and simply use the reward of a truncated state sequence; however, a value function would be straightforward to learn with TD-learning on optimized plans}. We summarize the main design choices below, deterministic LatCo in \cref{alg:plan}, and provide further details in \cref{app:platco}.

\begin{algorithm}
\caption{Latent Collocation (LatCo)}
\label{alg:plan}
    \begin{algorithmic}[1]
    \STATE Start with any available data $\mathcal{D}$
    \WHILE{not converged}
        \FOR{each time step $t~=~1 \dots T_\text{tot}$ \textbf{with step} $ T_\text{cache} $} 
        \STATE Infer latent state: $\lat_t \sim q(\lat_t | \obs_t)$
        \STATE Define the Lagrangian: 
            \begin{align}
            \begin{split}
             \mathcal{L}(\lat_{t+1:t+H},& a_{t:t+H}, \lambda)   = \sum_t \big[ r(\lat_t) \\ 
             - \lambda^{\text{dyn}}_t & (||\lat_{t+1} - \mu(\lat_t,a_t)||^2 - \epsilon^\text{dyn}) \\
             - \lambda^{\text{act}}_t & (\text{max}(0, |a_t| - a_{m})^2 - \epsilon^\text{act}) \big]. 
            \end{split}
            \end{align}
        \FOR{each optimization step $k~=~1 \dots K$} %
            \STATE Update plan: \\ $\lat_{t+1:t+H}, a_{t:t+H} \pluseq \tilde\nabla \mathcal{L} $ \COMMENT{Eq (\ref{eq:grad})}
            \STATE Update dual variables: \\ $\lambda_{t:t+H} := \textsc{Update}(\mathcal{L}, \lambda_{t:t+H}) $ \COMMENT{Eq (\ref{eq:lambda_upd})}
        \ENDFOR
        \STATE Execute $a_{t:t+T_\text{cache}}$ in environment: \\ $o_{t:t+T_\text{cache}},r_{t:t+T_\text{cache}} \sim p_{env}$
        \ENDFOR
        \STATE Add episode to replay buffer: \\$\mathcal{D} := \mathcal{D} \cup (o_{1:T_\text{tot}},a_{1:T_\text{tot}},r_{1:T_\text{tot}})$
    
        \FOR{training iteration $i~=~1 \dots \text{It}$} %
            \STATE Sample minibatch from replay buffer: \\ $(o_{1:T},a_{1:T},r_{1:T})_{1:b} \sim {\mathcal{D}}$
            \STATE Train dynamics model: \\$\phi \pluseq \alpha \nabla \mathcal{L}_\text{ELBO}(o_{1:T},a_{1:T},r_{1:T})_{1:b}$  \COMMENT{Eq (\ref{eq:elbo})}
        \ENDFOR
    \ENDWHILE
    \end{algorithmic}
\end{algorithm}

\textbf{Latent state models.} 
We implement the latent dynamics with convolutional neural networks for the encoder and decoder, and a recurrent neural network for the transition dynamics, following \citet{hafner2018learning,denton2018stochastic}. The latent state includes a stochastic component with conditional Gaussian transitions, and the hidden state of the recurrent neural network with deterministic transitions. Thanks to the latent MDP design, we never need to decode images during optimization, making it memory-efficient. 

\textbf{Dynamics relaxation.}
The dynamics constraint can be relaxed in the beginning of the optimization, leading LatCo to rapidly discover high-reward state space regions (potentially violating the dynamics), and then gradually modifying the trajectory to be more dynamically consistent, as illustrated in \cref{fig:teaser}. This is in contrast to shooting methods, which suffer from local optima in long horizon tasks, since the algorithm must simultaneously drive the states toward the high-reward regions \emph{and} discover the actions that will get them there. The ability to disentangle these two stages by first finding the high-reward region and only then optimizing for the actions that achieve that reward allows LatCo to solve more complex and temporally extended tasks while suffering less from local optima. In our approach, this dynamics relaxation is not explicitly enforced, but is simply a consequence of using primal-dual constrained optimization. %

\textbf{MPC and Online training.} LatCo is a planning method that can be used within an online model-based reinforcement learning training loop. In this setup, we use model predictive control (MPC) within a single episode, i.e. we carry out the plan only up to $T_\text{cache}$ actions, and then re-plan to provide closed-loop behavior. Further, for online training, we perform several gradient updates to the dynamics model after every collected episode. This provides us with a model-based RL agent that can be trained either from scratch by only collecting its own data, or by seeding the replay buffer with any available offline data.

\begin{figure*}
    \centering
    \includegraphics[width=1\linewidth]{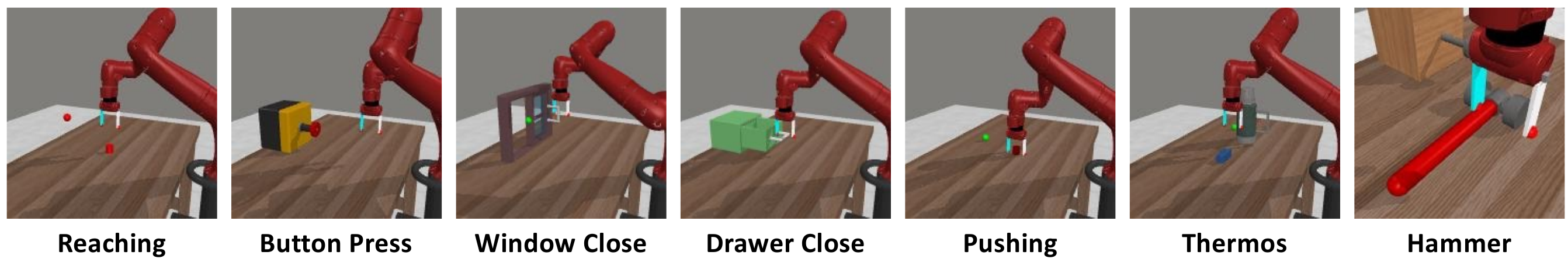}
    \vspace{-0.3in}
    \caption{Sparse MetaWorld tasks, featuring temporally extended planning and sparse rewards. The agent observes the environment only through the visual input shown here. LatCo creates effective visual plans and performs well on all tasks. }
    \label{fig:environments}
    \vspace{-0.1in}
\end{figure*}

\section{Optimization for Latent Collocation}
\label{sec:impl_details}

The latent collocation framework described in the previous section can be used with any optimization procedure, such as gradient descent or Adam \citep{kingma2014adam}. 
However, we found that the choice of the optimizer for both the latent states and the Lagrange multipliers has a large influence on runtime. We detail our specific implementation below.

\textbf{Levenberg-Marquardt optimization.} We use the Levenberg-Marquardt optimizer for the states and actions, which pre-conditions the gradient direction with the matrix $(\JT \J)^{-1}$, where $\J$ is the Jacobian of the objective with respect to states and actions.
This preconditioner approximates the Hessian inverse, %
significantly improving convergence speed:
\begin{equation}
    \label{eq:grad}
    \tilde\nabla = (\JT \J + \lambda I)^{-1} \JT \rho.
\end{equation}
The Levenberg-Marquardt optimizer has cubic complexity in the number of optimized dimensions. However, by noting that the Jacobian of the problem is block-tridiagonal, we can implement a more efficient optimizer that scales linearly with the planning horizon \citep{mordatch2012discovery}. This efficient optimizer converges 10-100 times faster than gradient descent in wall clock time in our experiments.

The Levenberg-Marquardt algorithm optimizes the sum of squares $\sum_i \rho_i^2$, defined in terms of residuals $\rho$.
Any bounded objective can be expressed in terms of residuals by a combination of shifting and square root operations. For the dynamics constraint, we use the $\lat_{t+1} - \mu(\lat_t,a_t)$ differences as residuals directly,
with one residual per state dimension. We similarly constrain the planned actions to be within the environment range $a_m$ using the residual $\max(0, |a_t| - a_{\text{m}})$. {This corresponds to using a squared constraint instead of a linear one}. For the reward objective, we form residuals as the softplus of the  negative reward: $\rho = \ln (1 + e^{-r}) $, which we found to be an effective way of forming a nonnegative cost without the need to estimate the maximum reward.

\textbf{Constrained optimization.} The naive gradient ascent update rule $\lambda \pluseq ||z_{t+1} - \mu(z_t, a_t)||^2 - \epsilon$ for the multiplier $\lambda$ works poorly when the current value of the multiplier is either much smaller or larger than the cost value. While it yields good plans, it suffers from slow convergence and suboptimal runtime. We can design a better behaved update rule by applying a monotonic function that rescales the magnitude of the update while preserving the sign, which can be seen as a time-dependent learning rate. 
Specifically, we observed that scaling the update with the value of the multiplier itself $\lambda$ as well as using log of the constrained value $\log ||z_{t+1} - \mu(z_t, a_t)||^2 - \log \epsilon$ provided better scaled updates and led to faster convergence:
\begin{equation}
    \label{eq:lambda_upd}
    \lambda \pluseq \alpha \log\left(\frac{||z_{t+1} - \mu(z_t, a_t)||^2}{\epsilon} + \eta \right) \lambda,
\end{equation}
where $\eta=0.01$ ensures numerical stability and the learning rate $\alpha = 0.1$. {Using a small non-zero $\epsilon$ is beneficial for the optimization and ensures fast convergence, as the exact constraint might be hard to reach.}

\section{Experiments}

We evaluate long-horizon planning capabilities of LatCo for model-based reinforcement learning on several challenging manipulation and locomotion tasks. Each subsection below corresponds to a distinct scientific question that we study.

\subsection{Experimental Setup}

\begin{table*}
\centering

\caption{MBRL results on the visual Sparse Metaworld tasks and DM Control. On sparse reward tasks, Shooting only solves the simpler tasks, while the powerful trajectory optimization with LatCo finds good trajectories more consistently.}
\vspace{-.1in}

\begin{footnotesize}
\begin{tabular}{llllll|lll}
\toprule

& Reaching & Button & Window & Drawer & Pushing & Reacher Easy & Cheetah Run & Quadruped Walk
\\

Shaped reward 
& {$\times$} 
& {$\times$} 
& {$\times$} 
& {$\times$} 
& {$\times$} 
& {\checkmark} 
& {\checkmark} 
& {\checkmark} 
\\

\midrule

LatCo (Ours)
&  \textbf{91 $\pm$ 3\%} &  \textbf{55 $\pm$ 4\%} &  \textbf{49 $\pm$ 8\%} &  \textbf{46 $\pm$ 3\%} &  \textbf{38 $\pm$ 3\%} &  559 $\pm$ 15 &  245 $\pm$ 12 &  \textbf{121 $\pm$ 9} \\

PlaNet
&  20 $\pm$ 1\% &  13 $\pm$ 2\% &  31 $\pm$ 2\% &  22 $\pm$ 2\% &  22 $\pm$ 3\% &  561 $\pm$ 14 &   \textbf{326 $\pm$ 3} &     72 $\pm$ 4 \\

MPPI
&  16 $\pm$ 1\% &  10 $\pm$ 2\% &  30 $\pm$ 2\% &  21 $\pm$ 3\% &  21 $\pm$ 4\% & 657 $\pm$ 17 &   298 $\pm$ 1 &     57 $\pm$ 4 \\

Shoot. GD
&  8 $\pm$ 1\% &   7 $\pm$ 0\% &  28 $\pm$ 1\% &  18 $\pm$ 2\% &  19 $\pm$ 3\% &   \textbf{756 $\pm$ 7} &   246 $\pm$ 2 &    101 $\pm$ 2 \\

\bottomrule
\end{tabular}
\label{tab:results_online}
\end{footnotesize}

\vspace{-0.2in}
\end{table*}

\begin{table}
\centering

\caption{MBRL with offline and online data. Shooting fails to construct adequate plans on these challenging long-horizon tasks, while LatCo performs significantly better.} 
\vspace{-.1in}

\begin{footnotesize}
\begin{tabular}{lll}
\toprule

& Thermos & Hammer
\\

Shaped reward 
& {$\times$} 
& {$\times$}
\\

\midrule

LatCo (Ours)
& \textbf{37 $\pm$ 21\%} &  \textbf{13 $\pm$ 2\%} \\

PlaNet
& 1 $\pm$ 1\% &   1 $\pm$ 0\% \\

MPPI 
& 0 $\pm$ 0\% &   0 $\pm$ 0\% \\

Shoot. GD
& 0 $\pm$ 0\% &   3 $\pm$ 1\% \\

\bottomrule
\end{tabular}
\label{tab:results_demos}
\end{footnotesize}

\vspace{-0.1in}
\end{table}

\textbf{Environments.}
To evaluate on challenging visual planning tasks, we adapt the MetaWorld benchmark~\citep{yu2020meta} to visual observations and sparse rewards, with a reward of 1 given only when the task is solved and 0 otherwise. \cref{fig:environments} shows example visual observations provided to the agent on these seven tasks. The robot and the object position are randomly initialized. The Thermos and Hammer tasks require a complex two-stage solution using tools to manipulate another object. We use $H = 30, T_\text{cache} = 30, T_\text{tot} = 150$ for all tasks except Pushing, Thermos, and Hammer, where we use $H = 50, T_\text{cache} = 25, T_\text{tot} = 150$. 
In addition, we evaluate on the standard continuous control tasks with shaped rewards from the DeepMind Control Suite \citep{tassa2020dmcontrol}. According to the protocol from \cite{hafner2019dream}, we use an action repeat of 2 and set $H = 12, T_\text{cache} = 6, T_\text{tot} = 1000$ for all DMC tasks.

\textbf{Comparisons.}
To specifically ascertain the benefits of collocation, we must determine whether they stem from gradient-based optimization, optimizing over states, or both. Therefore, we include a prior method based on zeroth-order CEM optimization, \textit{PlaNet} \citep{hafner2018learning}; another sampling-based shooting method, \textit{MPPI} \citep{williams2016aggressive,nagabandi2020deep};
as well as a gradient-based method that optimizes actions directly using the objective in \cref{eq:trajopt}, which we denote \textit{Shooting GD}. To isolate the effects of different planning methods, we use the same dynamics model architecture for all agents. \changed{We train all methods online according to \cref{alg:plan}.} The hyperparameters are detailed in \cref{sec:appendix-A,sec:appendix-B}. We use an action repeat of 2 for Thermos and Hammer environments for all methods, and no action repeat on other tasks.

\subsection{Is LatCo better able to solve sparse reward tasks in visual model-based reinforcement learning?}

First, we evaluate LatCo's performance in the standard model-based reinforcement learning setting, where the agent learns the model from scratch and collects the data online using the LatCo planner, according to \cref{alg:plan}. We evaluate the deterministic LatCo and other model-based agents on the Sparse MetaWorld tasks from visual input without any state information. %
The performance is shown in \cref{tab:results_online} and learning curves in the Appendix. We observe that LatCo is able to learn a good model and construct effective plans on these sparse reward tasks. Shooting trajectory optimization methods, however, struggle, not being able to optimize a sparse reward signal that requires longer-horizon planning.

\begin{figure}%
    \centering
    \vspace{-0.05in}
    \includegraphics[width=\linewidth]{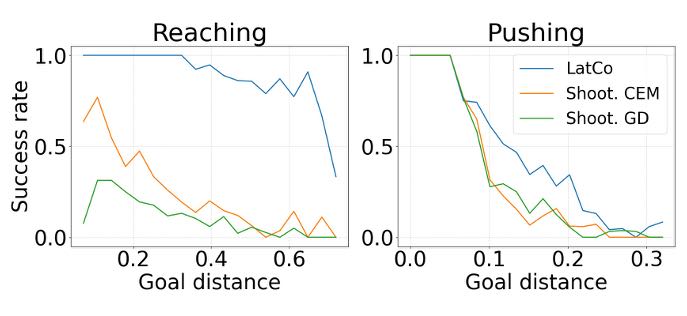}
    \vspace{-0.35in}
    \caption{Performance with respect to task difficulty. We observe that LatCo maintains good performance even for harder tasks with long-horizon goals, whereas shooting is only able to solve easier tasks.}
    \label{fig:parametric}
    \vspace{-0.2in}
\end{figure}

We further examine how the performance changes with the required planning horizon. We plot performance with respect to different distances to the goal in \cref{fig:parametric}. Shooting baselines degrade significantly on harder tasks that require longer horizons, while LatCo is able to solve even these harder tasks well. This confirms our hypothesis that collocation scales better to long-horizon tasks. 

In addition to the sparse reward tasks, we further evaluate our method on the more standard tasks from the DM control suite, also in \cref{tab:results_online} and dense MetaWorld tasks in App. \cref{fig:dense}. We use the experimental protocol from \cite{hafner2019dream} and our results are consistent with their Fig. 10. Since these tasks generally have dense rewards that are easy to optimize, we do not expect a significant improvement from LatCo. However, LatCo performs competitively with other methods and outperforms them on some environments, showcasing its generality.

\subsection{Is LatCo able to plan for and solve long-horizon robotic tasks from images?}
\label{sec:quantitative_exp}

\begin{figure*}%
    \centering
    \includegraphics[width=1\linewidth]{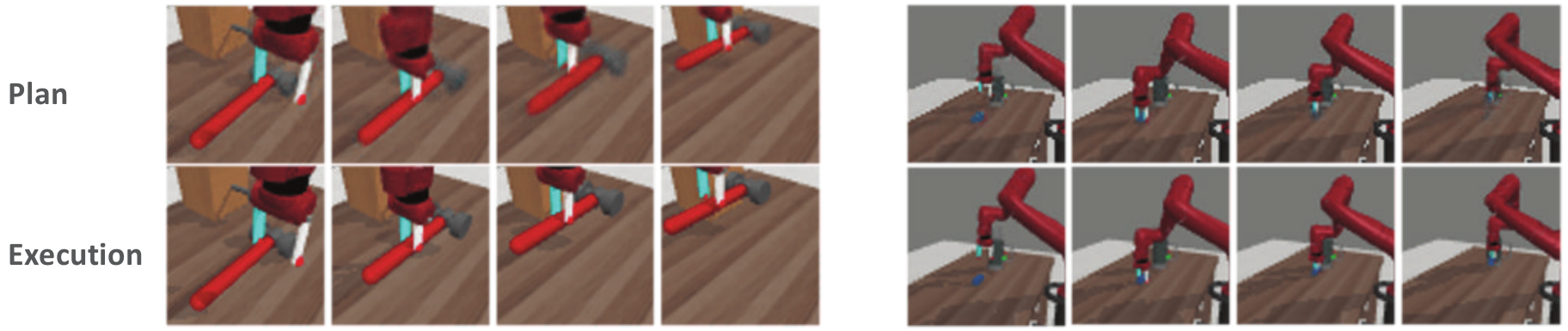}
    \vspace{-0.25in}
    \caption{
    Planned and executed trajectories on the hammer and thermos tasks. Plans are visualized by passing the latent states through the convolutional decoder. Both tasks require picking up a tool and using it to manipulate another object. The sparse reward is only given after completing the full task, and the planner needs to infer all stages required to solve the task. LatCo produces feasible and effective plans, and is able to execute them on these challenging long-horizon tasks.
    }
    \label{fig:strips}
    \vspace{-0.15in}
\end{figure*}

\begin{figure}%
    \centering
    \includegraphics[width=0.91\linewidth]{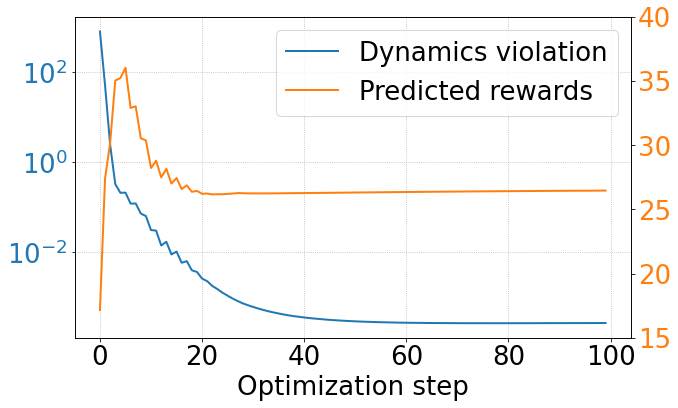}
    \vspace{-0.1in}
    \caption{Dynamics violation and reward of the plan over 100 optimization steps. First, LatCo explores high-reward regions, converging to the goal state. As dynamics are enforced, the plan becomes both feasible and effective.}
    \label{fig:optimization_curves}
    \vspace{-0.1in}
\end{figure}

Our aim is to evaluate LatCo on tasks with longer horizons than used in prior work (e.g., \citet{ebert2018visual,hafner2018learning}), focusing specifically on the performance of the planner. Solving such tasks from scratch also presents a challenging exploration problem, with all methods we tried failing to get non-zero reward, and is outside of the scope of this work.
Therefore, to focus on comparing planning methods, we use a pre-collected dataset with some successful and some failed trajectories to jump-start all agents, similarly to the protocol of \citet{Rajeswaran-RSS-18,nair2020accelerating}. We initialize the replay buffer with this dataset, and train according to \cref{alg:plan} for the Thermos and Hammer tasks shown in \cref{fig:environments}.
Further, during evaluation, we reinitialize the optimization several times for LatCo and Shooting GD and select the best solution. These optimization runs are performed in parallel and do not significantly impact runtime. We do not reinitialize CEM as it already incorporates sampling in the optimization, and it requires a large batch size, making parallel initializations infeasible.

From \cref{tab:results_demos}, we observe that LatCo exhibits superior performance to shooting-based methods on both long-horizon tasks. These tasks are challenging for all of the methods, as the dynamics are more complex and they require accurate long-term predictions to correctly optimize the sparse signal. Specifically, since there is no partial reward for grasping the object, the planner has to look ahead into the high reward final state and reason backwards that it needs to approach the object. We observed that both of these tasks require a planning horizon of at least 50 steps for good performance. %
Shooting methods struggle to find a good plan on these tasks, often stuck in the local optimum of not reaching for the object. LatCo outperforms shooting methods by a considerable margin, and is often able to construct and execute effective plans, as shown in \cref{fig:strips}.

\subsection{What is important for LatCo performance?}

\begin{table}
\centering

\caption{Ablating LatCo components on the Button task.} 
\vspace{-0.05in}

\begin{footnotesize}
\begin{tabular}{lll}
\toprule

&  Success & Runtime (FPS) \\

\midrule

LatCo
& 55\% & 1.6  \\

\midrule

LatCo no relaxation
& 38\% & 1.6  \\

LatCo no constrained opt.
& 40\% & 1.6  \\

LatCo no deterministic latent
& 43\% & 1.6  \\

LatCo no second-order opt.
& 48\% & 0.1  \\

Image Collocation
& 18\% & 0.2  \\

\bottomrule
\end{tabular}
\label{tab:ablations}
\end{footnotesize}

\vspace{-0.1in}
\end{table}

In this section, we analyze different design choices in LatCo.
First, we analyze the ability to temporarily violate dynamics in order to effectively plan for long-term reward. 
We show the dynamics violation costs and the predicted rewards over the course of optimization quantitatively in \cref{fig:optimization_curves} and qualitatively in \cref{fig:teaser}. Since the dynamics constraint is only gradually enforced with the increase of the Lagrange multipliers, the first few optimization steps allow dynamics violations in order to focus on discovering high rewards (steps 0 to 10 in \cref{fig:optimization_curves}). Later in the optimization, the constraints are optimized until the trajectory becomes dynamically feasible. If this ability to violate dynamics is removed by initializing the Lagrange multipliers $\lambda$ to large values, the optimization performs similarly to shooting and struggles to discover rewards, as shown in App. \cref{fig:additional_optimization_curves}.

We further evaluate the importance of various design choices quantitatively in \cref{tab:ablations} through ablations. We see that LatCo without dynamics relaxation performs poorly, confirming our qualitative analysis above. Using a constant balance weight instead of Lagrange multipliers (LatCo no constrained opt.) requires extensive tuning of the weight, but can perform well with the optimal value of the weight. This highlights the importance of our constrained optimization framework that removes the need for this additional tuning. Using a latent dynamics model with only probabilistic states (LatCo no deterministic latent) degrades the performance slightly as this architecture produces inferior predictions, consistent with the findings of \cite{hafner2018learning}. Using gradient descent instead of the Levenberg-Marquardt as well as for the Lagrange multiplier update (LatCo no second order) produces reasonable performance, but has much higher runtime complexity as it requires many more optimization steps. Optimizing images directly (Image Collocation) as opposed to optimizing latents performs better than shooting, but substantially worse than LatCo as the optimization problem is more complex.

\subsection{Does probabilistic LatCo handle uncertainty well?}

\label{sec:platco_exp}
Deterministic LatCo evaluated in the previous sections performs well in standard environments as they do not require reasoning about uncertainty. However, uncertainty is important for many types of control tasks \cite{kaelbling1998planning,thrun1999monte,kappen2005path}. In this section, we show that deterministic LatCo and common shooting methods fail a simple lottery task, unlike Gaussian LatCo (\cref{sec:platco}).

The Lottery task has two goals (\textit{Top} and \textit{Bottom}). Reaching the Top goal provides 1 reward. Reaching the Bottom goal provides either 20 ($65\%$ chance) or -40 reward ($35\%$ chance). The agent needs to navigate to the goal using continuous $x-y$ actions, and only one goal can be chosen. An agent that treats uncertainty incorrectly might plan to achieve 20 reward with the Bottom goal, which leads to an expected reward of -1. The correct solution is the Top goal.

\begin{figure}%
    \centering
    \includegraphics[width=1\linewidth]{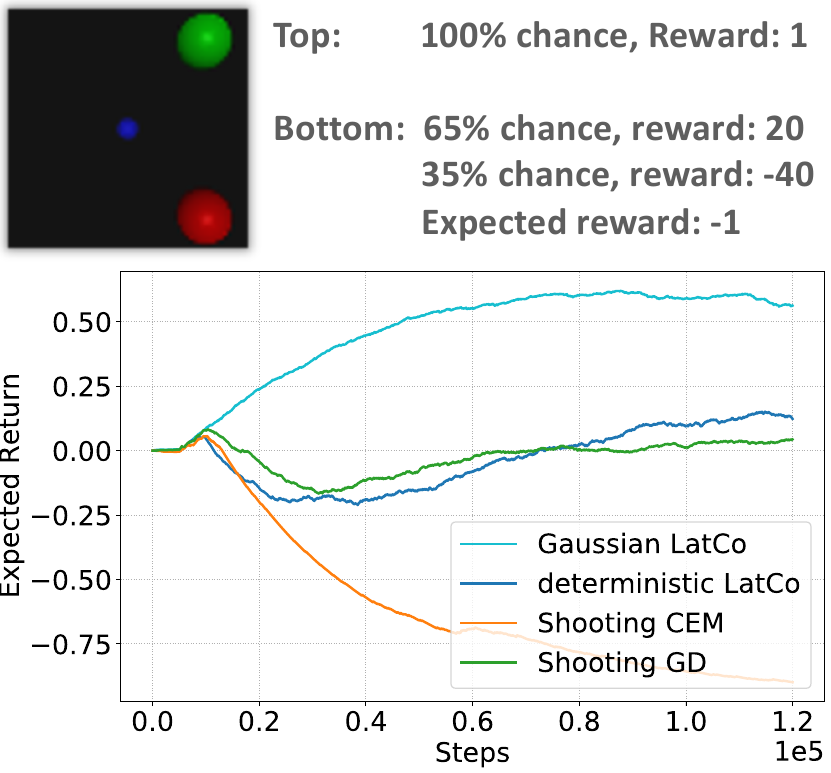}
    \vspace{-0.25in}
    \caption{Planning under uncertainty on the Lottery task. Top: the environment description, bottom: training curves. Gaussian LatCo is able to correctly plan in this environment, while deterministic latco and other shooting baselines fail as they are not designed to plan under uncertainty. LatCo correctly estimates the expected cumulative reward, accounting for both optimistic and pessimistic outcomes.}
    \label{fig:platco}
    \vspace{-0.2in}
\end{figure}

\cref{fig:platco} shows the performance of all methods, pretrained on a dataset of reaching either goal in equal proportion. As CEM chooses best sampled trajectories, it plans optimistically and prefers the incorrect Bottom goal. Deterministic LatCo and Shooting GD plan for the mean latent future, which is either optimistic or pessimistic depending on the latent space geometry. These methods visit both goals in similar proportions. Only Gaussian LatCo plans correctly by estimating the future state distribution preferring the Top goal. We further observed that while Gaussian LatCo only approximates the future using a Gaussian, this yields accurate prediction results in practice.

In \cref{fig:results_online,fig:dense} in the Appendix we show that Gaussian LatCo performs comparably to the deterministic version on MetaWorld tasks. In contrast to commonly used shooting methods, LatCo performs well in both stochastic and deterministic environments, further showcasing its generality.

\section{Discussion}

\textbf{Conclusion.} We presented LatCo, a method for latent collocation that improves performance of visual-model based reinforcement learning agents. In contrast to the common shooting methods, LatCo uses powerful trajectory optimization to plan for long-horizon tasks where prior work fails. By improving the planning capabilities and removing the need for reward shaping, LatCo can scale to complex tasks more easily and with less manual instrumentation. 

\textbf{Limitations.} 
As collocation usually requires many more optimization variables than shooting, it can be slower or take more optimization steps. While we address this with a problem-specific optimizer, future work might learn smooth latent spaces for easier planning. Further, we observed collocation can still suffer from local optima, although less so than shooting. This issue may be addressed with even larger overparameterized latent states or better optimization. 

\textbf{Future work.} By formulating a principled approach to optimizing sequences of latent states for RL, LatCo provides a useful building block for future work. Ideas presented in this paper may be applied to settings such as state-only imitation learning or hierarchical subgoal-based RL. With state optimization, state constraints in addition to rewards could be used for more flexible task specification.  As LatCo does not require sampling from the dynamics model, it allows the use of more powerful models such as autoregressive, energy-based or score-based models for which sampling is expensive. Finally, the benefits of state optimization may translate into the policy learning setting by learning an amortized policy to output future states with the LatCo objective.

\section*{Acknowledgements}
We thank Clark Zhang and Matt Halm for generous advice on trajectory optimization, and Aviral Kumar, Danijar Hafner, Dinesh Jayaraman, Michael Posa, Ed Hu, Suraj Nair, Karl Schmeckpeper, Russell Mendonca, Deepak Pathak, and members of GRASP for discussions, as well as anonymous reviewers for helpful comments and suggestions. Support was provided by the  ARL DCIST CRA W911NF-17-2-0181, ONR N00014-17-1-2093,  and by Honda Research Institute.

\bibliography{bibref_definitions_long,bibtex}
\bibliographystyle{icml2021}

\clearpage

\appendix
\section{Model Architecture and Training Details}
\label{sec:appendix-A}

\begin{figure*}%
    \centering
    \includegraphics[width=1\linewidth]{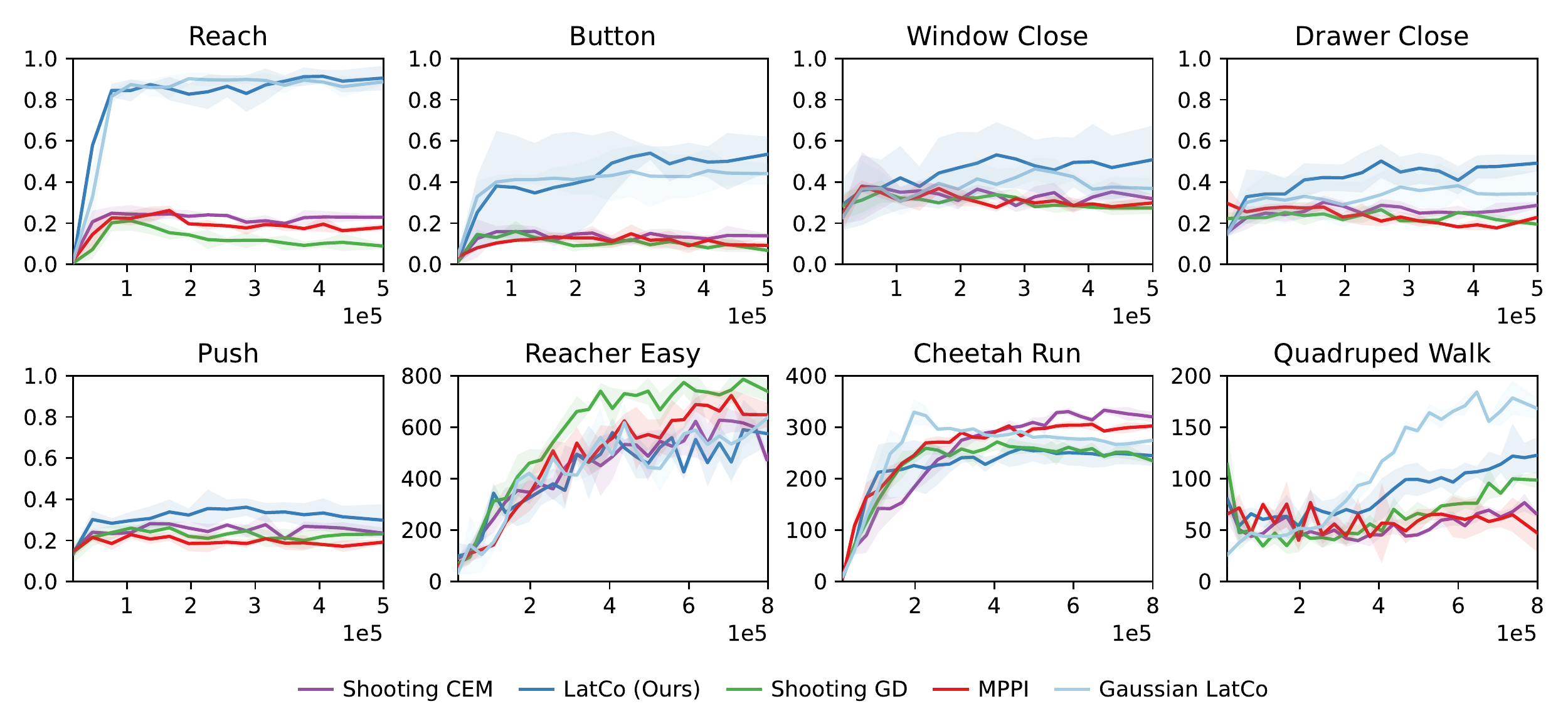}
    \vspace{-0.25in}
    \caption{Learning curves for online MBRL experiments. We also include a comparison against Gaussian LatCo. We observe that Gaussian LatCo performs comparably to the deterministic version. }
    \label{fig:results_online}
\end{figure*}

\begin{figure}%
    \centering
    \includegraphics[width=1\linewidth]{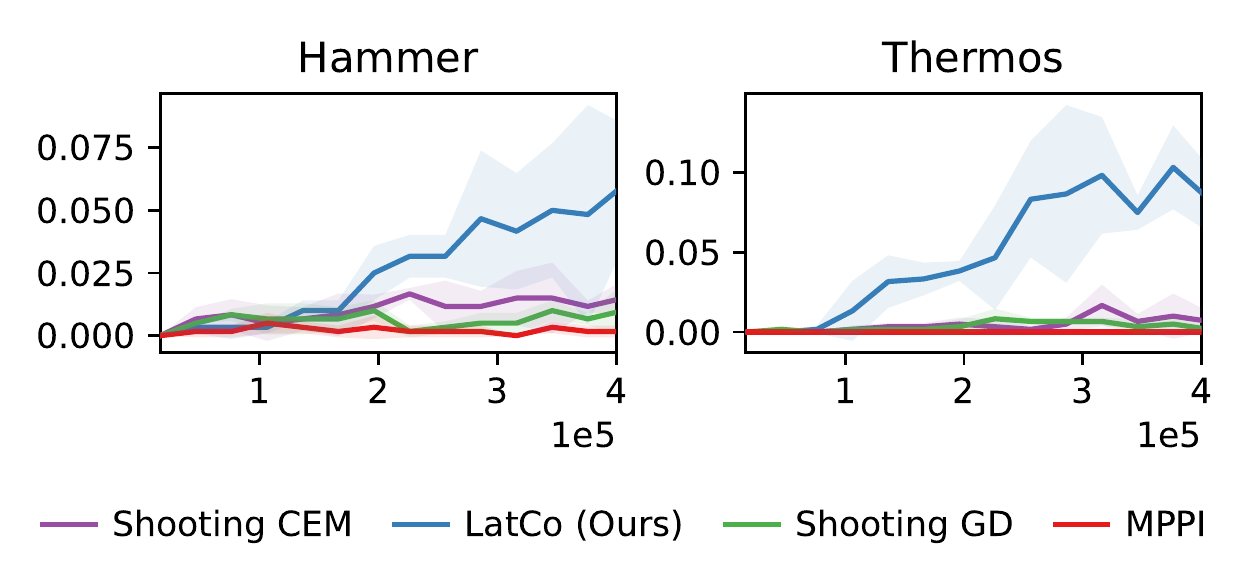}
    \vspace{-0.3in}
    \caption{Learning curves for MBRL experiments with offline and online data. This experiment is trained without parallel reinitializations. The results in \cref{tab:results_demos} are obtained from this trained agent, evaluated with parallel reinitializations. }
    \label{fig:results_demos}
    \vspace{-0.2in}
\end{figure}

We use the latent dynamics and reward models from PlaNet \citep{hafner2018learning} with default hyperparameters unless specified otherwise.
We set the image size to 64x64. For metaworld tasks, we train for $\text{It} = 15$ iterations for every $N = 1$ episode collected, while DMC uses the default hyperparameters. All models are trained on a single high-end GPU.

\section{Planning Details}

\begin{algorithm}
\caption{Gaussian LatCo}
\label{alg:prob_plan}
    \begin{algorithmic}[1]
    \STATE Start with any available data $\mathcal{D}$
    \WHILE{not converged}
        \FOR{each time step $t~=~1 \dots T_\text{tot}$ \textbf{with step} $ T_\text{cache} $} 
        \STATE Infer latent state: $\lat_t \sim q(\lat_t | \obs_t)$
        \STATE Define the Lagrangian: 
            \begin{align}
            \begin{split}
                \mathcal{L}(&\mu_{t+1:t+H},\sigma_{t+1:t+H},a_{t:t+H}, \lambda) = \sum_t \big[  \\ 
                &\E_{q(z_t)} \left[r(z_t)\right] \\
                &  - \lambda^{\text{dyn}}_t  (||\text{mean}[p(z_{t})] - \mu_t||^2 - \epsilon) \\
                &  - \lambda^{\text{dyn}}_t (||\text{stddev}[p(z_{t})] - \sigma_t||^2 - \epsilon)\\
                &  - \lambda^{\text{act}}_t  (\text{max}(0, |a_t| - a_{m})^2 - \epsilon^\text{act})\\
                \big]. 
            \end{split}
            \end{align}
        \FOR{each optimization step $k~=~1 \dots K$} %
            \STATE Update plan: \\ $\mu_{t+1:t+H},\sigma_{t+1:t+H}, a_{t:t+H} \pluseq \tilde\nabla \mathcal{L} $ \COMMENT{Eq (\ref{eq:grad})}
            \STATE Update dual variables: \\ $\lambda_{t:t+H} := \textsc{Update}(\mathcal{L}, \lambda_{t:t+H}) $ \COMMENT{Eq (\ref{eq:lambda_upd})}
        \ENDFOR
        \STATE Execute $a_{t:t+T_\text{cache}}$ in environment: \\ $o_{t:t+T_\text{cache}},r_{t:t+T_\text{cache}} \sim p_{env}$
        \ENDFOR
        \STATE Add episode to replay buffer: \\$\mathcal{D} := \mathcal{D} \cup (o_{1:T_\text{tot}},a_{1:T_\text{tot}},r_{1:T_\text{tot}})$
    
        \FOR{training iteration $i~=~1 \dots \text{It}$} %
            \STATE Sample minibatch from replay buffer: \\ $(o_{1:T},a_{1:T},r_{1:T})_{1:b} \sim {\mathcal{D}}$
            \STATE Train dynamics model: \\$\phi \pluseq \alpha \nabla \mathcal{L}_\text{ELBO}(o_{1:T},a_{1:T},r_{1:T})_{1:b}$  \COMMENT{Eq (\ref{eq:elbo})}
        \ENDFOR
    \ENDWHILE
    \end{algorithmic}
\end{algorithm}

\begin{algorithm}
\caption{Gaussian LatCo Lagrangian computation}
\label{alg:prob_lagrangian}
    \begin{algorithmic}[1]
    \STATE Given: initialized plan $\mu_{t+1:t+H},\sigma_{t+1:t+H},a_{t:t+H}$.
    \FOR{time $t$ within planning horizon}
        \STATE Sample $K=50$ latent states from the plan distribution: $z^k_t \sim q(z_t)$
        \STATE Evaluate the reward term with samples: $\E_{q(z_t)} r(z_t) \approx \sum_k r(z_t^k)$
        \STATE Approximate the one-step prediction distribution with samples: $p(z_t) \approx \{z^k_{t+1}\}_{k=1..K}$, where $z^k_{t+1} \sim p(z_{t+1} | z_t^k, a_t)$.
        \STATE Evaluate the sample mean of the one-step prediction distribution: $\text{mean}[p(z_{t+1})] \approx \frac{1}{K}\sum_k z^k_{t+1}$.
        \STATE Evaluate the sample standard deviation of the one-step prediction distribution: $\text{stddev}[p(z_{t+1})] \approx \sqrt{\frac{1}{K} \sum_k (z^k_{t+1} - \text{mean}[p(z_{t+1})])^2} $.
    \ENDFOR
    \STATE Define the Lagrangian: 
        \begin{align}
        \begin{split}
            \mathcal{L}(&\mu_{t+1:t+H},\sigma_{t+1:t+H},a_{t:t+H}, \lambda) = \sum_t \big[  \\ 
            &\E_{q(z_t)} \left[r(z_t)\right] \\
            &  - \lambda^{\text{dyn}}_t  (||\text{mean}[p(z_{t})] - \mu_t||^2 - \epsilon) \\
            &  - \lambda^{\text{dyn}}_t (||\text{stddev}[p(z_{t})] - \sigma_t||^2 - \epsilon)\\
            &  - \lambda^{\text{act}}_t  (\text{max}(0, |a_t| - a_{m})^2 - \epsilon^\text{act})\\
            \big]. 
        \end{split}
        \end{align}
    \end{algorithmic}
\end{algorithm}
\label{sec:appendix-B}

\paragraph{CEM \& MPPI.} On Metaworld tasks, we optimize for 100 iterations, where in each iteration, 10000 action sequences are sampled and the distribution is refit to the 100 best samples.  On DM Control tasks, we optimize for 10 steps, where in each iteration, 1000 action sequences are sampled and the distribution is refit to the 100 best samples. The MPPI shares most hyperparameters with CEM, with the additional parameter $\gamma = 10$. For MPPI, we observed that both 10000 and 1000 action sequences yield similar results, so we use 1000 in our final results.  {We have manually tuned these hyperparameters, and we report the best results.}

\paragraph{GD.} On Metaworld tasks, we optimize for 500 iterations using the Adam optimizer \citep{kingma2014adam} (which is a modified version of momentum gradient descent) with a learning rate 0.05. We use dual descent to penalize infeasible action predictions. The Lagrange multipliers are updated every 5 optimization steps. On DM Control tasks, all hyperparameters are the same except that we optimize for only 100 iterations. {We have manually tuned the learning rate and tried several first-order optimizers, and we report the best results.}

\paragraph{LatCo.} On Metaworld tasks, we optimize for 200 iterations using the Levenberg-Marquardt optimizer with damping $10^{-3}$. {The damping parameter controls the trust region, with smaller or zero damping speeding up convergence, but potentially leading to numerical instability or divergence.} The Lagrange multipliers are updated every step using the rule from Section \ref{sec:impl_details}, with $\epsilon^\text{dyn} = \epsilon^\text{act} = 10^{-4}$ and $\eta = 0.01$. {The threshold $\epsilon$ directly controls the magnitude of the final dynamics and action violations. In general, we found this parameter to be most important for good performance, as a large threshold may cause infeasible plans, while a low threshold would make the initial relaxation of the dynamics constraint less effective. We observed that a single threshold of $10^{-4}$ works for all of our Metaworld environments. $\eta$ controls the update of the Lagrange multipliers. A larger $\eta$ makes the optimization more aggressive but less stable, and a smaller $\eta$ diminishes the effect of multiplier updates. We initilalize $\lambda^\text{dyn}_0 = 1, \lambda^\text{act}_0 = 1$.} On DM Control tasks, all hyperparamers are the same except that we optimize for 100 iterations and set $\epsilon^\text{dyn} = \epsilon^\text{act} = 10^{-2}$.

\paragraph{Gaussian LatCo.} The hyperparameters for Gaussian LatCo are largely the same as for deterministic LatCo. However, we observed that Gaussian LatCo requires less optimization steps to converge and use 50 iterations. Further, since the threshold $\epsilon^\text{dyn}$ is now used for both the mean and the variance, it is set to $10^{-2}$

\paragraph{LatCo no relaxation.} We prevent the relaxation of dynamics constraint by initializing the Lagrange multipliers to $10^8$. 

\paragraph{LatCo no constrained optimization.} We manually tune the multiplier values and fix them to $\lambda^\text{dyn} = 8, \lambda^\text{act} = 16$ throughout optimization.

\paragraph{LatCo no second order.} We use 5000 optimization steps and update the Lagrange multipliers every 5 steps instead of 1. Further, for this ablation we use the simple gradient descent update rule for lagrange multipliers with a learning rate of 1.5.

\paragraph{Image Collocation.} We use the same hyperparameters as the \textit{no second-order opt} ablation, except the learning rate, which is set to $0.02$.

\begin{figure*}%
    \centering
    \includegraphics[width=1\linewidth]{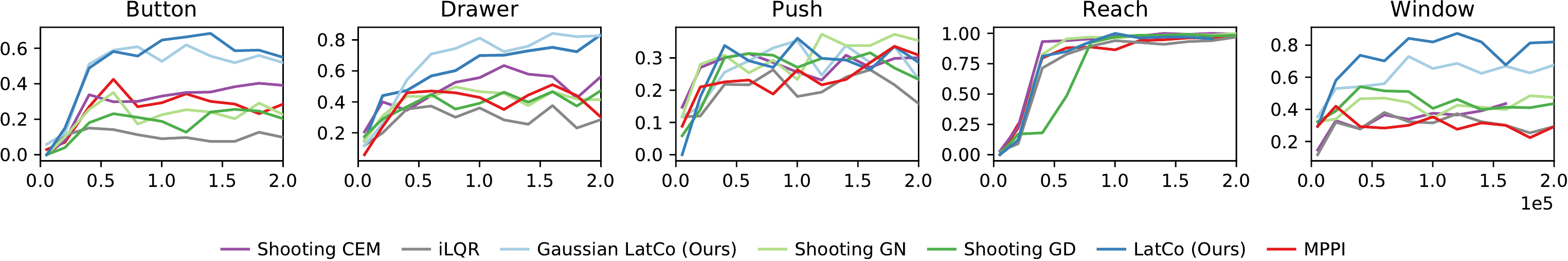}
    \vspace{-0.25in}
    \caption{Online MBRL results on the Metaworld tasks with dense rewards. With dense reward, collocation still outperforms shooting methods, however, shooting methods show more progress, especially on the reaching task. We include an additional iLQR baseline implemented with our latent dynamics model. This method performs poorly, likely due to the extremely non-linear and high-dimensional latent space. Further, we include a comparison against Gaussian LatCo. We observe that Gaussian LatCo performs comparably to the deterministic version. 
    }
    \label{fig:dense}
\end{figure*}

These planning hyperparameters remain fixed across the experiments as we observe that reward optimization converges in all cases. Planning a 30-step trajectory takes 12, 14, and 14 seconds for CEM, GD, and LatCo respectively on a single RTX 2080Ti graphics card. The action limits $a_m$ are set to the limits of the environment, in our case always $a_m = 1$.

 \begin{figure*}
    \centering
    \includegraphics[width=0.32\textwidth]{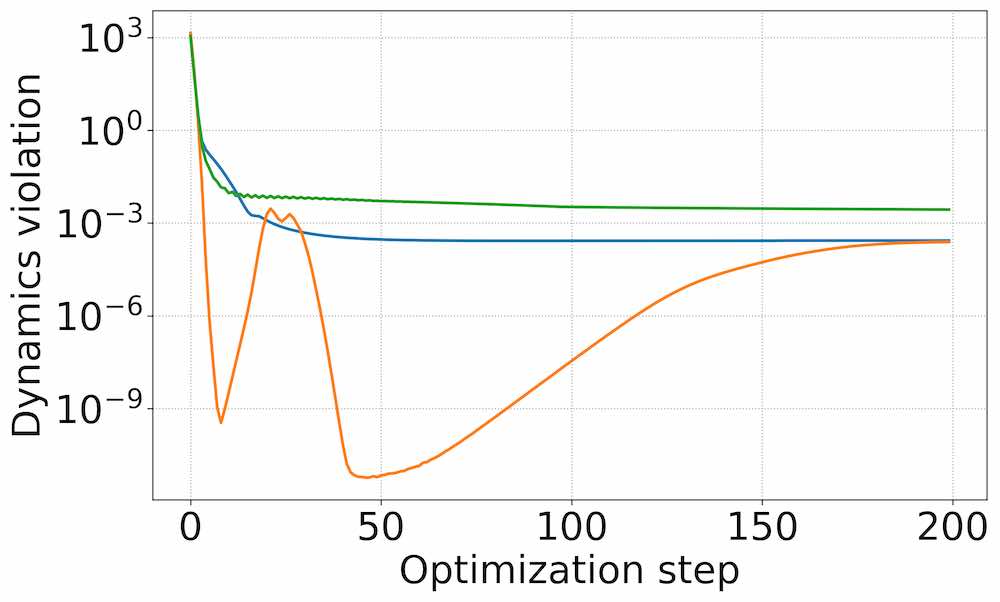}
    \includegraphics[width=0.32\textwidth]{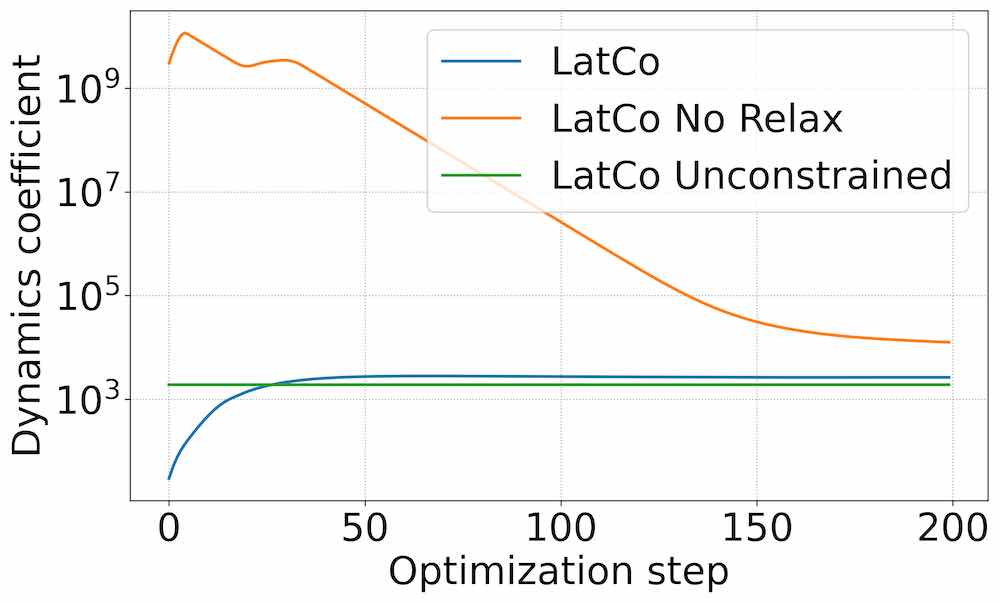}
    \includegraphics[width=0.32\textwidth]{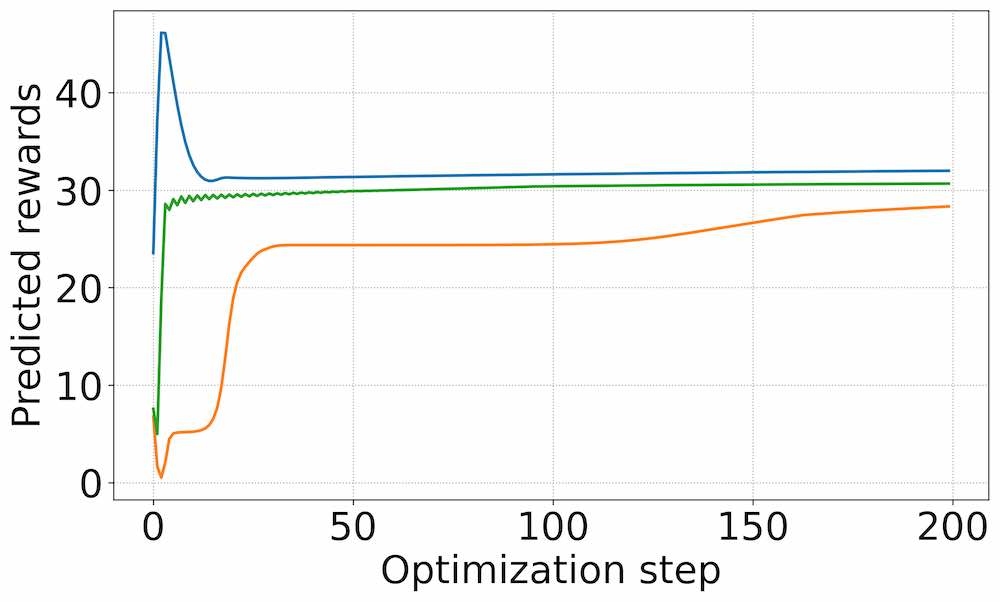}
    \caption{{Additional optimization curves. The dynamics coefficient (magnitude of Lagrange multipliers) increases exponentially as the dynamics constraint is enforced, and eventually converges} for LatCo. LatCo No Relax  does not relax the dynamics initially as the coefficient is initialized at a high value. We see that this leads to suboptimal reward optimization as the found solution is not as good as with the full LatCo. }
    \label{fig:additional_optimization_curves}
\end{figure*}

\section{Gaussian LatCo details}
\label{app:platco}

We provide a detailed algorithm for Gausssian LatCo in \cref{alg:prob_plan} and a detailed algorithm for evaluating the Lagrangian in \cref{alg:prob_lagrangian}. The gradients are estimated with reparametrization. We observed that optimization stability is improved by only optimizing the variance $\sigma$ with next step gradients. That is, for each variance parameter $\sigma_t$ we only optimize it with respect to terms of the Lagrangian that also involve $q(z_{t-1})$, but we set the gradient of any term that involves $q(z_{t+1})$ to zero. This is similar to the approach of \citet{patil2015scaling} and corresponds to a more shooting-like formulation since the variance information is only propagated forward in the trajectory and not backward. Even though this is not the full gradient of the Lagrangian, we observed optimization still works well as the variance can always flexibly match its target.

\section{Compared Methods and Ablations Details}

\textbf{CEM.} The Cross-Entropy Method (CEM) is a sampling-based trajectory optimization technique. The method maintains a distribution of action sequence, initialized as an isotropic unit Gaussian. In each iteration, the method samples $n$ action sequences from the current distribution and forwarded them through the model to obtain their predicted rewards. It then re-fits the distribution to the top $k$ trajectories with the highest rewards by computing their empirical mean and standard deviation.

\textbf{MPPI.} MPPI is a variant of CEM where instead of taking the average of the elite samples, it takes the weighted average when estimating the mean and variance of the new distribution. The weights are computed by taking the Softmax of the rewards with a temperature parameter $\gamma$. The temperature parameter effectively controls the width of the distribution.

\textbf{Shooting GD.} This method optimizes a single action trajectory. The trajectory is initialized from a uniform distribution and optimized with gradient descent, with the reward as the objective.  
 
\textbf{Shooting GN.} This method is similar to Shooting GD, but uses the Gauss-Newton (GN) optimizer instead of gradient descent. The Gauss-Newton optimizer in this case is easy to implement and is fast since the number of optimized variables is small for shooting.

\textbf{iLQR.} We additionally implemented an iLQR baseline with the same latent dynamics model as used in our method. We largely follow \citet{tassa2012synthesis} for our implementation. Specifically, we implement line search on the optimization step size, and multiply the trust-region regularizer by 2 when the $Q_{uu}$ matrix fails to be positive definite. We observed that using feedback controllers leads to worse results and only used the actions planned during optimization.

\textbf{Image collocation.} We directly optimize images instead of latent states using the same RSSM model. Images do not constitute a Markov space, therefore we optimize them recurrently. Specifically, we evaluate the reward of image sequence by encoding them into the latent space and predicting the reward. We evaluate the dynamics constraint on the frame $I_{t+1}$ by encoding the past sequence $I_{1:t}$ into the latent space, rolling out a one-step prediction, and decoding the images. Further, we observed poor optimization performance when optimizing recurrent constraints. Instead, we treat all constraints as pairwise by setting the gradients on $I_{1:t-1}$ to zero. We use gradient descent to optimize Image Collocation.

\section{Additional experimental results}
\label{sec:additional_exp}

\begin{figure}%
    \centering
    \includegraphics[width=\linewidth]{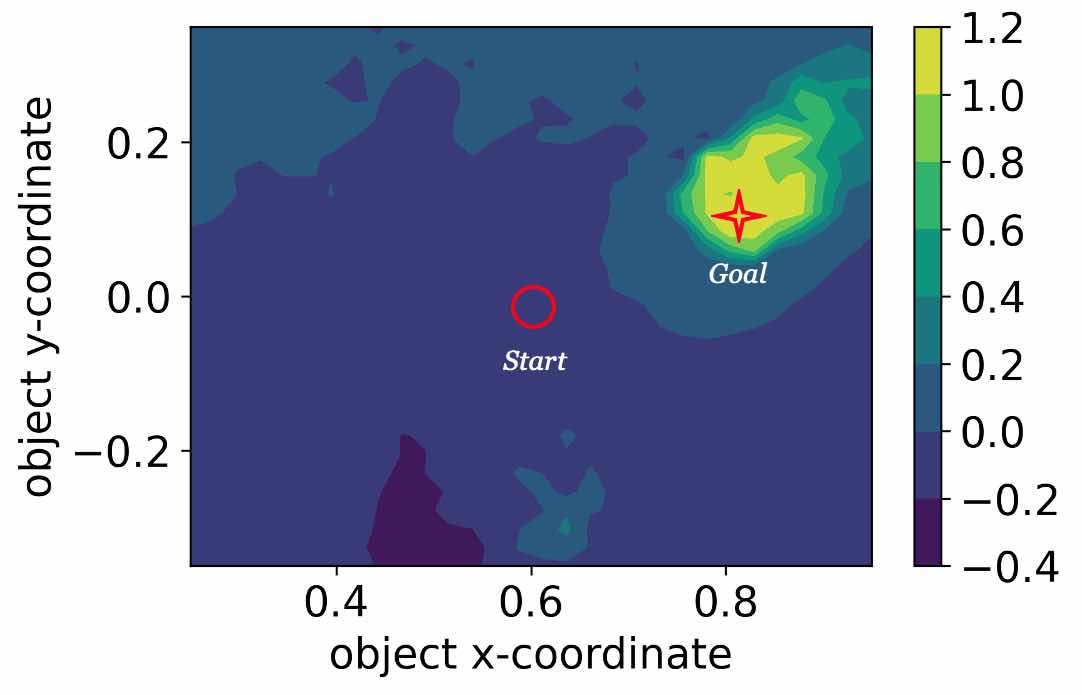}
    \caption{{Visualization of the reward predictor for the Sawyer Pushing task. The output of the reward predictor is shown for each object position on the 2D table. We see that the reward predictor correctly predicts a value of 1 at the goal, and low values otherwise. In addition, there is a certain amount of smoothing induced by the reward predictor, which creates a gradient from the start to the goal position. This explains why gradient-based planning is applicable even in this sparse reward task. We note that this reward smoothing is caused simply by the fact we are training a neural reward predictor, and does not require any additional setup.} 
    }
    \label{fig:reward_vis}
    \vspace{-0.2in}
\end{figure}

\subsection{Learning Curves}
We provide the learning curves for results in \cref{fig:results_online,fig:results_demos}. The online agents were trained for 500K environment steps on the MetaWorld tasks and 800K environment steps on the DM Control tasks. The hybrid agents were trained for 400K environment steps.

\subsection{Dense MetaWorld tasks and Gaussian LatCo results}
\label{sec:dense_exp}

We additionally evaluated all methods on the MetaWorld tasks with their original shaped rewards. We observed that the softplus reward transformation used by LatCo tends to squash large positive rewards, resulting in ineffective optimization of the reward objective. Thus, we keep the running mean and standard deviation of rewards during training, and normalize the environment rewards with these metrics. This keeps the reward magnitude in a reasonable range and facilitates LatCo training. Learning curves for the dense reward tasks are shown in Fig. \ref{fig:dense}.

\subsection{Analysing optimization}

We visualize the additional optimization curves in Fig \ref{fig:additional_optimization_curves}.

\subsection{Analysing sparse reward planning}
 
We observed that our method is able to solve sparse reward tasks using gradient-based planning. While this may be surprising at first, similar observations were made by prior work \cite{singh2019end}. We visualize the reward predictor output on the Pushing task in Fig \ref{fig:reward_vis}.

\section{Discussion of state optimization objectives}

\label{app:objectives}

We discussed graph-based methods for planning state trajectories in Sections 1 and 2. Here, we discuss in more detail other objectives for collocation, alternative to the constrained optimization objective proposed in this paper.

\paragraph{Maximum likelihood.} It is appealing to formulate the problem of planning future states as a maximum likelihood problem, such as the problem of finding the most likely trajectory given the start and the goal \cite{rhinehart2019precog,du2019model,pertch2020long}, or the most likely trajectory given a high reward \cite{janner2021reinforcement}. This casts the control problem in a probabilistic inference framework \cite{levine2018reinforcement}, for which many effective tools are already available. However, the solution recovered by this framework is known to be optimistic and does not correspond to the optimal control solution \cite{levine2018reinforcement,eysenbach2019if}. In a simple example, the best plan for playing the lottery is to win the lottery; however, trying to execute this plan would often lead to underwhelming outcomes and very low (indeed negative) expected reward. A similar phenomenon is demonstrated in our simple lottery task in \cref{sec:platco_exp}.

\paragraph{Goal reaching.} Another common alternative to constrained optimization is to formulate a goal-reaching objective, such as simple MSE between two observations \cite{nair2019hierarchical} or the amount of time needed to reach state B from state A \cite{pong2018temporal,nasiriany2019planning}. In this case, a sequence of states is optimized with such an objective pairwise, with the first state being the current state and the last state being the goal state. This is appealing as this framework naturally supports hierarchical planning if the goal-reaching objective is applicable to temporally far apart states. However, it also suffers from incorrectly treating future state probabilities. Indeed, it is unclear how to constrain the future states to be dynamically feasible in the latter approach. \citet{nasiriany2019planning} attempts to do so by adding a likelihood objective from a learned generative model, but this combined objective requires tuning a balance term between prior and reward and does not account for dynamics.

In practice, both above objectives are not probabilistically sound and require tuning a prior term that attempts to ensure the trajectory is physical. In contrast, our method does not require this tuning because it finds the right Lagrange multiplier automatically. Our method can be viewed as a generalization of the above approaches with the balance weight set automatically instead of manual tuning.

\end{document}